\documentclass[conference]{IEEEtran}
\IEEEoverridecommandlockouts
\usepackage{cite}
\usepackage{amsmath,amssymb,amsfonts}
\usepackage{xcolor}
\usepackage{algorithmic}
\usepackage{graphicx}
\usepackage{textcomp}
\usepackage{xcolor}
\usepackage[most]{tcolorbox}
\usepackage{tabularx}
\usepackage{subcaption}
\usepackage{float}
\usepackage{booktabs}   
\usepackage{amssymb}    
\usepackage{xcolor}     
\usepackage{fancyhdr}   
\usepackage{makecell} 

\newcommand{\cmark}{\textcolor{green!70!black}{\checkmark}}
\newcommand{\xmark}{\textcolor{red}{\texttimes}}
\IEEEoverridecommandlockouts
\usepackage{cite}
\usepackage{amsmath,amssymb,amsfonts}
\usepackage{algorithmic}
\usepackage{graphicx}
\usepackage{textcomp} 
\usepackage{xcolor}
\def\BibTeX{{\rm B\kern-.05em{\sc i\kern-.025em b}\kern-.08em
    T\kern-.1667em\lower.7ex\hbox{E}\kern-.125emX}}

\makeatletter
\newcommand{\linebreakand}{%
  \end{@IEEEauthorhalign}
  \hfill\mbox{}\par
  \mbox{}\hfill\begin{@IEEEauthorhalign}
}
\makeatother
\def\BibTeX{{\rm B\kern-.05em{\sc i\kern-.025em b}\kern-.08em
    T\kern-.1667em\lower.7ex\hbox{E}\kern-.125emX}}

\usepackage{framed}

\definecolor{buyergreen}{RGB}{0,51,102}      
\definecolor{sellerpurple}{RGB}{139,0,0}    

\definecolor{neutralgray}{RGB}{100,100,100}

\begin{document}

\title{Emotionally-Aware Agents for Dispute Resolution\\

}

\author{\IEEEauthorblockN{Sushrita Rakshit}
\IEEEauthorblockA{
\textit{University of Michigan}\\
Ann Arbor, MI \\
sushrita@umich.edu}
\and
\IEEEauthorblockN{James Hale}
\IEEEauthorblockA{
\textit{University of Southern Calirofnia}\\
Los Angeles, CA \\
jahale@usc.edu}
\and
\IEEEauthorblockN{Kushal Chawla}
\IEEEauthorblockA{\textit{Capital One} \\
New York City, NY \\
kushal.chawla@capitalone.com}
\linebreakand
\IEEEauthorblockN{Jeanne M. Brett}
\IEEEauthorblockA{
\textit{Northwestern University}\\
Evanston, IL \\
jmbrett@kellogg.northwestern.edu}
\and
\IEEEauthorblockN{Jonathan Gratch}
\IEEEauthorblockA{
\textit{University of Southern California}\\
Los Angeles, CA \\
gratch@ict.usc.edu}
}

\maketitle

\fancypagestyle{firstpage}{%
  \fancyhf{} 
  \renewcommand{\headrulewidth}{0pt} 
  \renewcommand{\footrulewidth}{0pt} 
}

\thispagestyle{firstpage}

\begin{abstract}
In conflict, people use emotional expressions to shape their counterparts' thoughts, feelings, and actions.  This paper explores whether automatic text emotion recognition offers insight into this influence in the context of dispute resolution. Prior work has shown the promise of such methods in negotiations; however, disputes evoke stronger emotions and different social processes. We use a large corpus of buyer-seller dispute dialogues to investigate how emotional expressions shape subjective and objective outcomes. We further demonstrate that large-language models yield considerably greater explanatory power than previous methods for emotion intensity annotation and better match the decisions of human annotators. Findings support existing theoretical models for how emotional expressions contribute to conflict escalation and resolution and suggest that agent-based systems could be useful in managing disputes by recognizing and potentially mitigating emotional escalation.
\end{abstract}

\begin{IEEEkeywords}
Dispute, Mediation, Affective Computing, Emotional Modeling, Emotion Recognition, Human-Agent-Interaction
\end{IEEEkeywords}

\section{Introduction}

Emotional expressions serve essential social functions in human relationships. They convey one's beliefs, desires, and intentions --- shaping the beliefs, desires, and intentions of interaction partners~\cite{parkinson2005emotion,scarantino2017things}. People high in emotional intelligence achieve more success in navigating emotional relationships~\cite{jordan2021managing}, and there exists growing interest in creating AI agents that understand and enact these social functions~\cite{celsoFunction23,jung2017affective}. Prior work suggests that emotionally-aware agents are suitable for a growing list of applications, including teaching people to convey emotions effectively~\cite{foster2016using}, improving human-agent interaction~\cite{stock2022survey}, detecting and moderating toxic communication~\cite{d2020bert}, and serving as methodological tools for studying human emotion~\cite{kennedy2023moral}. 

This paper examines the capacity of agents to understand human emotional expressions in the context of text-based dispute resolution. Disputes arise when one party in a relationship (an individual, group, or nation) levies a claim that another party refuses to accept, thus threatening the future of the relationship~\cite{felstiner2017emergence}. Though related to negotiation, which the AI community has studied extensively~\cite{baarslag2017will, faratin1998negotiation, kraus1997negotiation, jonker2012negotiating, aydougan2020challenges,gratch2015negotiation}, disputes involve unique social processes~\cite{brett2014negotiating} and have received less attention. Negotiation (or deal-making) involves coming together to create a new relationship (i.e., focus on opportunities for gains), and when parties can't reach a deal, they can seek other partners --- e.g., if they cannot deal with one car dealer, one can always visit another.
In contrast, disputes involve an existing relationship that has gone badly. As parties are already linked, success depends on managing the costs of ending the relationship rather than on opportunities moving forward. As a result, disputes evoke much stronger emotions than negotiations, particularly anger. Unlike negotiation, where expressions of anger promote compromise~\cite{van2004interpersonal,de2011effect}, anger provokes retaliation in a dispute~\cite{pruitt2007conflict}. Disputants become entrenched in their positions and seek to overpower their opponent through appeals to justice (``You violated my rights!'') or threatening harm (``I will sue you!''), leading to a spiral of escalation, sometimes including threats of physical violence~\cite{brett1998breaking, halperin2008group,pruitt2007conflict}. Thus, the costs of disputes can greatly exceed the original perceived injury, engulfing not only the disputants but also other related parties and even society at large. Agents who understand emotional dynamics in disputes and how to diffuse destructive behavior could have an enormous societal impact. 

Our work makes several contributions to emotion-aware agents. Prior research on emotions in text focuses on a document as the unit of analysis and outputs coarse-grained ``sentiment'' rather than specific emotions (e.g., \textit{anger}) \cite{wankhade2022survey}. In contrast, the dispute literature shows how specific emotions convey particular intentions and examines how these conveyed intentions change \textit{within} a text. This presents a challenge as interpreting each utterance depends on the context of prior utterances ~\cite{poria2019emotion}. To our knowledge, emotion recognition has not yet been systematically studied in dispute resolution. In negotiations, emotion recognition has shown only weak promise to predict outcomes --- e.g., Chawla \textit{et al.}~\cite{chawla2023towards} find that recognized \textit{anger} predicted satisfaction in the final agreement, but only explained 5\% of the variance. Though two observations suggest we can do much better in disputes. First, as emotions, like \textit{anger}, play a much larger role in disputes than negotiations, we hypothesize that emotion recognition will yield significantly more insights into disputes relative to negotiations. Second, recent research in large language models suggests they are remarkably adept at reasoning about human emotion, including the ability to incorporate contextual information~\cite{tak2023acii,yongsatianchot2023investigating,zhan2023evaluating,broekens2023fine,wang2023emotional}. Thus, the current paper assesses the ability of LLMs --- specifically GPT4o, but we also survey others --- to annotate emotion intensities in dispute dialogues; using these intensities in a linear model to predict disputants' outcome satisfaction; and yielding insight into how emotional expressions shape these outcomes. To foreshadow our findings, we find that automatically recognized emotional expressions can, in some cases, explain over 40\% of the variance in dispute outcomes (compared with about 5\% in prior negotiation research). This shows the importance of emotions in disputes and implies the possibility of agent-based methods that understand and influence human disputes --- e.g., mediation. 

For the current study, we examine a large corpus of 2,025 text conversations where online participants act as buyers or sellers in a simulated purchase dispute. The dispute was crafted in collaboration with a dispute resolution expert (Jeanne M. Brett) to evoke strong emotions while adhering to the ethical guidelines for human experimentation. The corpus was collected using the same framework as Chawla et al.'s \cite{chawla2021casino} CaSiNo --- specifically, we created the dialogue collection application using Lioness Labs \cite{giamattei2020lioness}. This allows participants to match online and engage in a dispute via text chat. Prior analysis of negotiations collected with CaSiNo found that emotional expressions, recognized by a large fine-tuned model (Google's T5), could predict participants' satisfaction with their negotiated agreement~\cite{chawla2023towards}. Thus, our paper seeks to extend these findings to disputes and contrast T5 with more recent models (out-of-the-box LLMs). 

This paper reports our success in predicting the consequences of emotional expressions in disputes. First, after validating our approach against self-reports, we show that automatically recognized expressions predict subjective outcomes --- i.e., each participant's feelings about the dispute's outcome. 
Secondly, we illustrate how textual emotion recognition gives insight into how disputes unfold. Our results replicate prior social science findings on the role of \textit{anger} in escalation and lay a foundation for AI agents that could detect escalation and intervene before a dispute derails.
Further, while social science findings have emphasized the role of \textit{anger} in disputes, our findings highlight the importance of additional emotional expressions, such as \textit{compassion}.
Finally, we survey T5 against popular LLMs predicting subjective outcomes and matching human annotators.

\section{Related Work}


Disputes and negotiations are task-oriented, mixed-motive interactions. They are task-oriented as each party comes to the situation with specific goals --- thus, disputes have concrete measures of success (i.e., ``What goals were achieved?''). They are mixed-motive because these goals, while not necessarily zero-sum, typically misalign --- i.e., parties can find better solutions if they exchange information about each other's interests and find solutions to maximize joint gains~\cite{thompson1991information,baarslag2016learning}.

AI researchers have long worked on non-collaborative task-oriented interactions, mostly focusing on negotiation framed as bilateral (occasionally multiparty) multi-issue bargaining~\cite{fatima2004agenda}:  parties are concerned with multiple issues that combine to form an overarching (private) objective function they seek to optimize. Parties reach agreements by exchanging offers and possibly other information.  Most research examines agents that negotiate with other agents, but in recent years, there has been growth in agents that negotiate with people. Human-agent negotiation has focused on structured menu-based communication (e.g., ~\cite{mell2016iago}), but LLMs have sparked increased interest in text-chat negotiations~\cite{lewis2017deal,he-etal-2018-decoupling,chawla-etal-2023-selfish}. Our work brings AI methods to the study of text-chat human-human disputes. 

Though disputes share many elements with negotiations, they are less studied than negotiations, as they involve argumentation over facts; most negotiating agents only exchange offers and do not allow explanations or justifications. 
An exception is research on argument-based negotiation~\cite{parsons1998agents}, though that line of work traditionally focuses on agent-agent interaction. Other work on disputes has focused on mediation wherein parties are prevented from interacting with each other but rather communicate to an impartial mediator that proposes fair solutions (e.g., \cite{goldman2015spliddit}). Neither line of work considers direct interactions between people, nor do they consider the role of emotional expressions. Thus, we focus on emotional expression in human disputes as a contribution of our work.

Expressions of emotion play an important role in shaping negotiations and disputes~\cite{morris2000emotions,douglas2014attitude} though relatively little work has used automated methods to investigate emotional expressions, and most of this work has focused on negotiation. Most work uses simple dictionary-based approaches~\cite{laubert2019you}, though more recent work takes advantage of powerful transformer-based models~\cite{chawla2023towards}. A crucial consideration in emotion research is the selection of labels. Emotion recognition has largely focused on sentiment (positive vs negative) or the six ``basic emotion'' labels proposed by the emotion psychologist Paul Ekman ~\cite{ekman1992argument}. Within affective science, there is a growing critique of these basic emotions for recognizing how people truly feel~\cite{barrett2019emotional}; however, they serve as a common vocabulary for labeling the expressions of others, and much of the social science work on emotion in disputes uses these terms. 
Thus, this paper adopts basic emotional labels as a starting point to allow direct comparison with prior work, with minor alterations justified later on, while acknowledging that other schemes may be more appropriate, particularly in cross-cultural contexts where common emotion terms may have differing meanings~\cite{havaldar2023multilingual}. 


\begin{figure*}[h]
\centering
\begin{framed}
\small 
\renewcommand{\arraystretch}{1.25} 
\begin{tabular}{@{}p{1.1cm}@{\hspace{0.35em}}p{0.85\linewidth}@{}}



\textbf{\textcolor{buyergreen}{Buyer:}}  & \textit{I am seeking a refund for the order I received that was different from the item I requested.} \\

\textbf{\textcolor{sellerpurple}{Seller:}} & \textit{Sorry, I sent you what you ordered. I will give you \$50 if you return it. I will drop my bad review if you drop yours, and I want an apology.} \\

\textbf{\textcolor{buyergreen}{Buyer:}}  & \textit{I disagree. I ordered a Kobe Bryant jersey, and that is not what I got. I will happily return the item for a full refund. I will remove the bad review if you remove yours of me, and I will not apologize because I did nothing wrong.} \\

\textbf{\textcolor{sellerpurple}{Seller:}} & \textit{I never offered a Kobe Bryant jersey --- I never had one, and that jersey would have cost at least \$3900.} \\

\textbf{\textcolor{buyergreen}{Buyer:}}  & \textit{I have a screenshot of the webpage on your site that I ordered the jersey from, and it clearly states that the item was a Kobe Bryant jersey for \$75.00.} \\

\textbf{\textcolor{sellerpurple}{Seller:}} & \textit{Now you are lying. I am a dealer. I know what that jersey is worth. If you do not agree to my terms, I will update your bad review.} \\

\textbf{\textcolor{buyergreen}{Buyer:}}  & \textit{You are wrong to accuse me of lying. I do have the screenshot and can provide you with a copy if you wish, or I can simply post it alongside my review of your business.} \\

\textbf{\textcolor{sellerpurple}{Seller:}} & \textit{If you do that, I will report you for traud. On the web, anyone can create untrue posts.} \\

\textbf{\textcolor{buyergreen}{Buyer:}}  & \textit{I do not know what ``traud'' is supposed to mean in this context, and apparently you do not intend to do the right thing. I will be forced to report this to the Attorney General and the Consumer Protection Bureau.} \\

\textbf{\textcolor{sellerpurple}{Seller:}} & \textit{I Walk Away.} \\

\end{tabular}
\end{framed}

\caption{A \texttt{KODIS} dialogue snippet illustrating how an initial focus on interests escalates into appeals to moral norms and ultimately threats of retaliation, resulting in an impasse.}
\label{fig:dialogexample}

\end{figure*}

While our work focuses on recognizing and analyzing emotion in conflict, other research highlights how agents could intervene or assist disputing humans. Cho et al.~\cite{cho-etal-2024-language} evaluated conversational AI that intervenes in emotional online disputes --- finding prompting models using insights from social science provided specific and fair feedback on toxic behavior. Other studies used AI agents to teach dispute-resolution skills.  Murawski \textit{et al.} developed an agent-based framework to teach family caregivers how to manage disputes with insurance companies and service providers through AI role-play~\cite{murawski2024negotiage}. The Rehearsal system extends this approach using large-language models (LLMs) to simulate emotion-laden interactions~\cite{shaikh2024rehearsal}. Rehearsal recognizes when statements likely escalate the conflict and provides personalized feedback on more cooperative dialogue moves --- results suggest the system improves students' dispute resolution abilities. 
The results of our work may serve to improve such systems.

\section{Dispute Resolution Corpus}
We analyze KODIS \cite{hale2025kodis}, a corpus of human dyadic disputes collected in English on Prolific.
Potentially identifiable information from participants was removed. The University of Southern California's Institutional Review Board approved the experimental design and classified it as minimal risk. Participants gave informed consent and could opt to leave the process at any point without penalty. The entire task, including pre- and post-questionnaires, took approximately twenty minutes. Participants were compensated a base rate of \$3.50 USD (\$10.50 per hour), plus a bonus of up to \$3 USD depending on how well they achieved their objectives. 

\subsection{Scenario}\label{AA}
The role-playing scenario involves a dispute over purchasing a Kobe Bryant basketball jersey. Before the task, each side (buyer and seller) received unique role-playing instructions designed to provoke a dispute and argumentation over facts. The buyer read that they purchased the jersey for their sick nephew, and when the wrong item arrived, the seller claimed it was never advertised as a Kobe jersey and refused a refund. The seller read that the 
buyer bought a generic Jersey and now complains about the item. Lastly, each side posted negative reviews about the other, accusing them of deception.
 In sum, the disputants discuss whether the seller should grant a refund, each side should drop their review, and each side should apologize. If they cannot agree on a resolution, they can choose to walk away. See Fig.~\ref{fig:dialogexample} for an example dialogue.
 

\begin{table*}[!tbh]
\centering

\small
\renewcommand{\arraystretch}{1.5}
    \begin{tabularx}{\textwidth}{l>{\raggedleft\arraybackslash}X>{\raggedleft\arraybackslash}X}
\textbf{Sample Utterances}                                                & \textbf{T5-Twitter} & \textbf{GPT4o} \\ \hline
I have proof, and you're not willing to look at it.                          & Joy                 & Anger           \\ \hline
But are you sure you will refund me completely?                            & Joy                 & Fear            \\ \hline
What if the shirt has been damaged?                                        & Joy                 & Fear            \\ \hline
If I may return the goods, will I get my money back?? Or I just lost it? & Sadness             & Fear            \\ \hline
You want to keep the item with no refund?                                 & Anger               & Surprise        \\ \hline
Huh?                                                                      & Joy                 & Surprise        \\ \hline
Oh, I'm sorry to hear that, what shirt did you received?                       & Sadness               & Compassion        \\ \hline
Hello. I need you to make this right. This is for my nephew that's sick. & Joy & Fear
\end{tabularx}
\caption{ Utterances with T5-Twitter and GPT Labels.  We portray the top-score associated label from GPT's intensity vector.}
\label{tab:examplesT5GPT}

\end{table*}
\subsection{Measures}
Participants complete several scales assessing their experience during the dispute. This includes a 10-item scale \cite{aslani2016dignity} about tactics used during the dispute ---e.g., ``I expressed frustration,'' ``The OTHER PARTY expressed frustration.'' Here, we only consider the 2-item \textit{frustration} sub-scale, averaging each side of the dyad to quantify the frustration expressed in the dialogue. As subjective perceptions of dispute are a better predictor of future negotiation decisions than the objective result~\cite{brown2012utility}, participants completed the Subjective Value Inventory (SVI) scale \cite{curhan2006people}, measuring four dimensions of subjective value. This includes feelings about the instrumental outcome, feelings about the process (\textit{Was it fair?}), feelings about the relationship with the partner, and feelings about the self (\textit{Did I lose face?}).

\subsection{Corpus Overview}
The 2,025 disputes\footnote{Gender breakdown of 50\% Female, 49\% Male, and 1\% other.} consist of 21,681 dialogue turns, averaging 10.7 turns per dispute. Of these, 19\% failed to reach a resolution --- despite a potentially sizeable monetary bonus. This contrasts with Chawla et al.'s \cite{chawla2021casino} CaSiNo negotiation corpus, where only 3\% of negotiations ended in an impasse. 
Consistent with prior research, this suggests greater impasse rates for disputes relative to negotiations.
Further, examining self-reported responses, we find strong evidence that expressed emotion influenced the disputes. For example, although 19\% of participants failed to reach an agreement, this climbs to 33\% in the dialogues in the top quartile of self-reported frustration and drops to only 6\% in the dialogues with the lowest.

\section{Improving Emotion Recognition}

Though prior work has demonstrated the salience of emotions in shaping disputes, automatically recognizing emotions in an unfolding conversation remains a challenge --- e.g., annotating the emotion of each utterance depends on the context of prior utterances ~\cite{poria2019emotion}. Prior work on negotiation has shown limited success in predicting outcomes from recognized expression, but this work predated the advances afforded by LLMs. 
In this section, we outline our proposed approach to leverage LLMs as automatic emotion recognition tools; though, first, we overview our selected baseline. 

\subsection{Baseline: T5}
As a baseline comparison against large language models, we adopt the best-performing approach used by Chawla et al. \cite{chawla2023towards} in their negotiation research --- the T5 model finetuned for emotion on the Twitter corpus. This pre-trained model classified utterances as one of \textit{joy}, \textit{anger}, \textit{love}, \textit{sadness}, \textit{fear}, or \textit{surprise}; the model classified each utterance of the dispute in isolation, and created an intensity vector using one-hot encoding. We adopt it for this dispute setting, as other researchers have popularized its use in conflict research --- specifically negotiation \cite{chawla2023towards, kwon2024llmseffectivenegotiatorssystematic}. 



One concern relates to the choice of emotion labels provided by T5-Twitter. Its labels arguably improve over Ekman's more common six basic labels, as T5-Twitter distinguishes positive emotions (\textit{love} vs. \textit{joy}), whereas Ekman \cite{ekman1992argument} only differentiates amongst negative emotions. That said, expressions of \textit{love} are rare in intense disputes. Instead, the conflict literature suggests that expressions of \textit{compassion} are a powerful predictor of negotiated outcomes~\cite{Allred97compassion,TingToomey14-compassion}. This indicates that an alternative labeling scheme might improve results, an adaptation that general LLMs easily allow.


\subsection{LLM-Based Emotion Recognition}\label{sec:LLMIntro}
We propose an alternative approach utilizing LLMs and explore several prompting strategies. 
In preliminary efforts, we attempted to apply annotation methods developed for deal-making dialogues in this dispute setting, though these underperformed, highlighting the need to treat disputes as a distinct genre.
We limit our initial reported analysis to OpenAI's GPT series of models ~\cite{achiam2023gpt} (GPT4o, run on 06/28/2024) as a recent extensive evaluation found GPT-4 yielded the best performance compared with a wide range of other LLMs on making inferences from negotiation dialogues~\cite{kwon2024llmseffectivenegotiatorssystematic}. Later, we survey additional LLMs to gauge relative performance. Motivated by prior findings~\cite{zheng2023helpful}, we assign a specific role in the prompt (``You are a good emotion classification tool''), and prompt the model to output instructions in JSON output format. Further, we leverage the following prompting strategies:

\subsubsection{Incorporating Dialogue Context}  
Given that emotions of each dialogue turn depend on prior turns~\cite{poria2019emotion}, we include dialogue history when prompting the LLM to generate emotion annotations. 
\subsubsection{Labeling Scheme} 
To address concerns with T5-Twitter's labels, we substituted T5-Twitter's \textit{love} with \textit{compassion} and added a \textit{neutral} label to allow annotation of emotionless utterances. We prompt the LLM to generate soft labels for each utterance --- i.e., it allocates weight ($\in[0,1]$) over \textit{joy}, \textit{anger}, \textit{fear}, \textit{surprise}, \textit{compassion}, \textit{sadness}, and \textit{neutral} labels such that the resulting vector sums to one. 

\subsubsection{In-context Learning} 
Recent research suggests performance can be improved via ``in-context learning'' \cite{dong2022survey}. Thus, we include in the prompt several dialogue turns from the corpus --- hand-annotated with emotion labels by the researchers. 

\section{Analyzing the Dispute Dialogues}
We examine whether automatic emotion recognition approaches can offer insights into the corpus in three steps. 
First, we contrast how well T5-Twitter and our proposed LLM prompts recognize expressed emotion compared against disputants' self-reports. Second, we assess how well these automatic labels can predict disputants' subjective feelings about the outcome of the dispute. Third, we examine how emotions unfold during the dispute, and if differences in emotional expression can indicate outcomes in advance. 

\subsection {Emotion Recognition Performance Validity}\label{sec:construct}
We first analyze how well GPT and T5's labels --- see TABLE~\ref{tab:examplesT5GPT} for comparative annotations --- match disputants' actual emotions ($N=1,938$).
Without the self-reports of each emotion, we assess how well T5 and GPT4o's emotion intensities\footnote{As T5 selects one emotion, we convert that to one-hot encoding.}, averaged over the entire dialogue, correlate with self-reported \textit{frustration}. GPT4o outperforms T5-Twitter at capturing the magnitude and direction of the correlation coefficients --- likely because T5-Twitter fails to annotate instances of \textit{fear} and \textit{sadness} accurately, giving more consideration to \textit{joy} and \textit{anger}. Examining the Pearson correlation coefficients in Table~\ref{frustration-emotion}, we see that \textit{fear}, as annotated by GPT4o, yields a coefficient of 0.360; while T5-twitter's, however, stands at a low 0.026. Notably, we see a much stronger correlation between GPT's \textit{anger} annotations and \textit{frustration} compared to T5. As T5-Twitter cannot capture a variety of labels --- such as \textit{fear} and \textit{sadness} --- GPT4o better captures nuanced emotional trends that match self-reports. 

\begin{table}[h!]
\small
\centering
\begin{tabular}{|l|c|c|}
\hline
\multicolumn{1}{|c|}{\textbf{Emotion}} & \textbf{\begin{tabular}[c]{@{}c@{}}T5-Twitter w/\\ Frustration\end{tabular}} & \textbf{\begin{tabular}[c]{@{}c@{}}GPT4o w/\\ Frustration\end{tabular}} \\ \hline
Anger                                  & 0.294                                                                     & 0.544                                                                 \\
Fear                                   & 0.026                                                                    & 0.360                                                                 \\
Sadness                                & 0.053                                                                     & 0.162                                                                 \\
Surprise                               & 0.013                                                                    & -0.010                                                                \\
Compassion                             & -                                                                         & -0.181                                                                \\
Love                                   & -0.024                                                                    & -                                                                     \\
Joy                                    & -0.326                                                                    & -0.357                                                                \\ \hline
\end{tabular}
\caption{Pearson correlation coefficients of emotion labels from T5-Twitter and GPT4o with self-reported \textit{frustration}.}
\label{frustration-emotion}
\end{table}

\begin{figure}[H]
  \centering
  \includegraphics[width=0.49\textwidth]{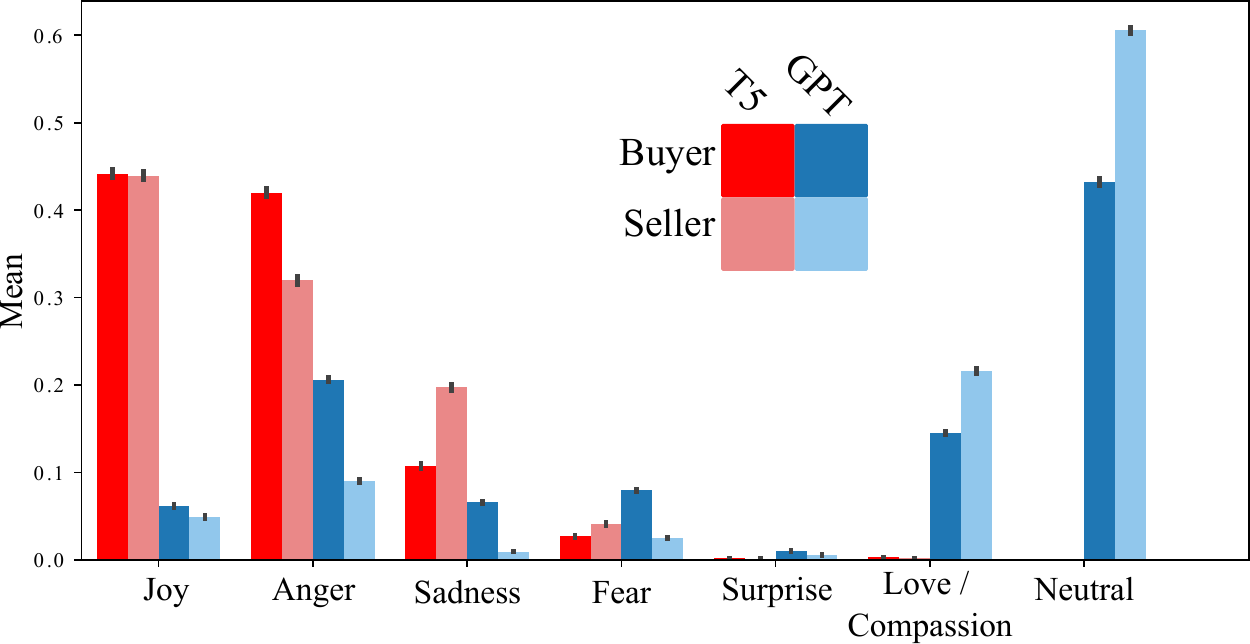} 
  \caption{Comparison of mean intensity of recognized emotions by GPT4o and T5-Twitter separated by role. 
  } 
  \label{fig:gpt_avgs} 
\end{figure}

To explain this, we explore differences in mean emotion broken out by role between GPT and T5-Twitter. Fig.~\ref{fig:gpt_avgs} shows that T5 skews towards annotating utterances as \textit{joy} or \textit{anger}, while GPT assigns more diverse labels and utilizes \textit{neutral} as a dampener. GPT suggests dialogues contain far more \textit{anger} than \textit{joy}, especially for \textit{buyers}. Interestingly, GPT also recognizes more \textit{compassion} relative to T5's \textit{love}. 

\subsection{Predicting Subjective Outcomes}\label{sec:em-comparison}
We next predict each side's subjective feelings about the dispute's outcome --- as measured by the SVI scale (Instrumental Outcome, Process Fairness, Relationship, and Self)~\cite{curhan2006people} --- from the automatically recognized emotions during the dispute. To justify our prompting approach, we first contrast T5-Twitter and the various prompting techniques --- ablating or modifying various aspects of the prompt --- to see which method's annotations better explain the subjective outcomes.
We perform these tests on a 20\% subset ($N=406$) of the corpus. 
We use multiple linear regression (MLR) to predict subjective feelings for each role, regressing each of the four SVI sub-scales on the average emotion intensities from a given dialogue. The coefficient of determination ($R^2$) measures model fit --- i.e., how much of the variance the emotion labels explain of the outcome. 

\begin{table}[ht]
\centering
\small
\renewcommand{\arraystretch}{1.3}
\begin{tabular}{lccccc}
\toprule
\textbf{Model} & \textbf{IC} & \textbf{w/} & \textbf{Dialogue} & \textbf{w/} & $R^2$ \\
& \textbf{Learning} & \textbf{Compassion} & \textbf{History} & \textbf{Neutral} & \\
\midrule
T5      & \xmark & \xmark & \xmark & \xmark & 0.112 \\
GPT4o  & \cmark & \xmark & \xmark & \xmark & 0.250 \\
GPT4o  & \cmark & \cmark & \xmark & \cmark & 0.258 \\
GPT4o  & \cmark & \cmark & \cmark & \cmark & \textbf{0.267} \\
\bottomrule
\end{tabular}
\caption{Mean $R^2$ for predicting SVI from emotion labels for different schemes.}
\label{tab:r2_results}
\end{table}

\begin{figure}[]
    \centering
    \includegraphics[width=.7\linewidth]{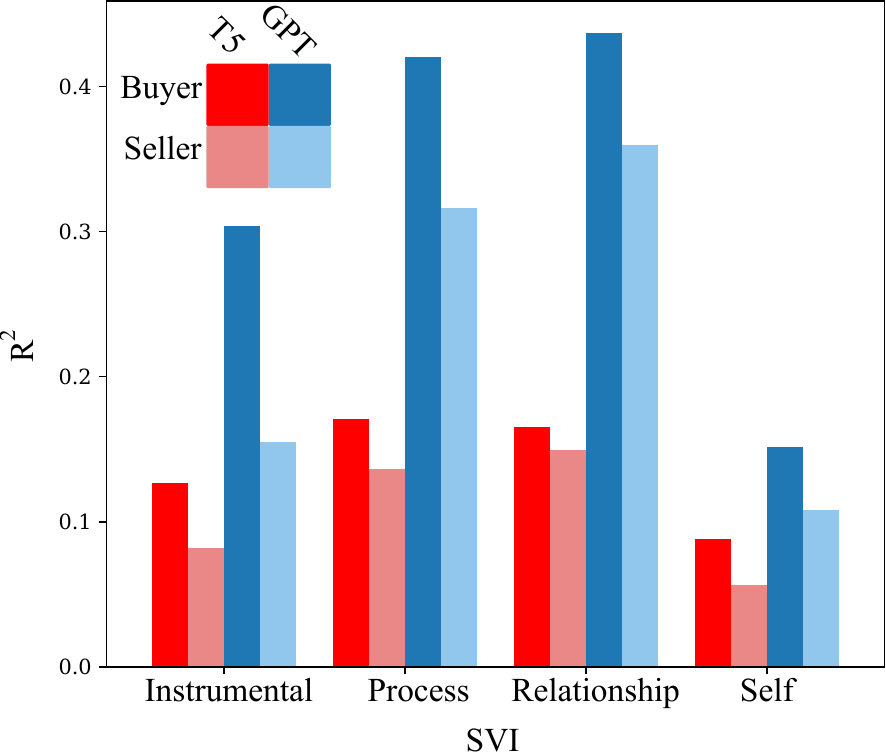}
    \caption{$R^2$ values for various regressions measuring the efficacy of the emotion labels to predict SVI scores.}
    \label{fig:SVI_MLR}.
\end{figure}

We compare T5 with different combinations of prompt enhancements for GPT4o, considering the average model fit ($R^2$) across all four SVI sub-scales --- see TABLE~\ref{tab:r2_results}. 
This shows that simply replacing T5 with GPT4o yields the largest improvement. Additional modest improvements come from adopting each of our proposed prompting methods from Section~\ref{sec:LLMIntro}. 
Fig.~\ref{fig:SVI_MLR} compares this highest-performing GPT configuration with T5 when breaking out the SVI sub-scales, and separating predictions by role in the dispute. This shows \textit{buyers} are more straightforward to predict: GPT explains almost half the variance in outcome about \textit{buyers'} feelings about the \textit{process} and \textit{relationship}. Now that we have justified our approach, we can begin to analyze the emotional dynamics in \texttt{KODIS}.

\begin{figure*}[h!]
  \centering
  \begin{subfigure}[b]{0.49\linewidth}
    \centering
    \includegraphics[width=\linewidth]{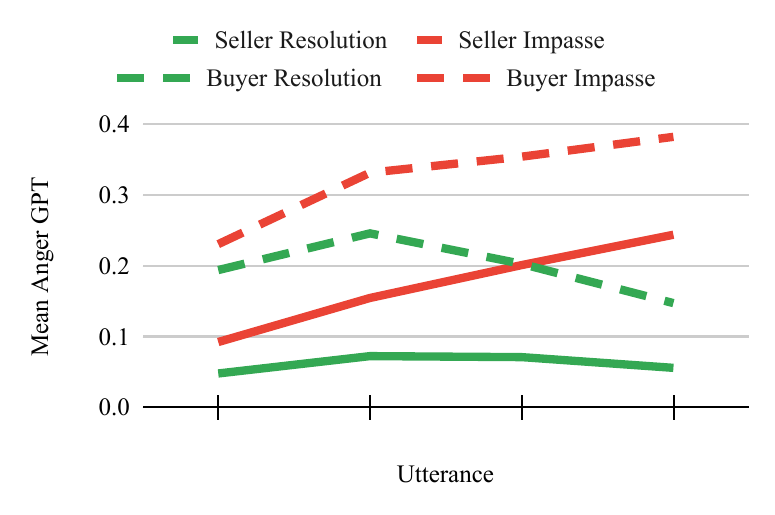} 
    \caption{Anger over time.}
    \label{fig:anger}
  \end{subfigure}
  \hfill
  \begin{subfigure}[b]{0.49\linewidth}
    \centering
    \includegraphics[width=\linewidth]{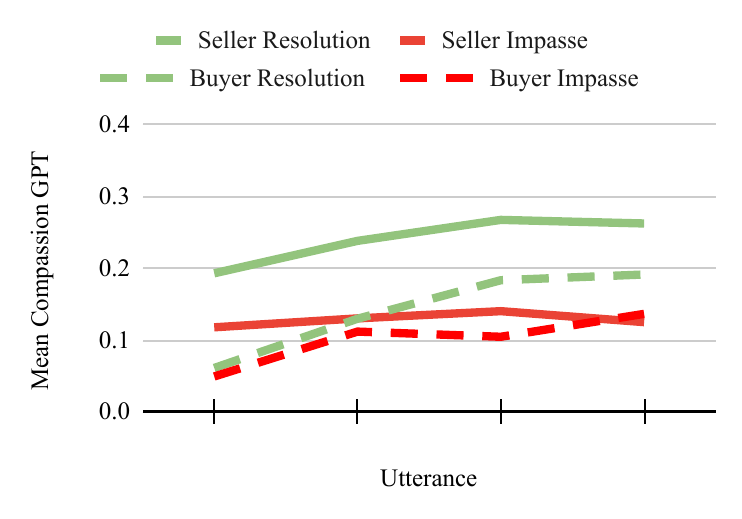} 
    \caption{Compassion over time.}
    \label{fig:compassion}
  \end{subfigure}
  \caption{Average expressed \textit{anger} and compassion across dialogue turns broken out by role and objective outcome.}
  \label{fig:emotiontime}
\end{figure*}

\subsection{Exploring Escalatory Spirals}
We find that outcomes can be predicted quite well solely from emotions expressed throughout the dispute --- in some cases explaining almost half of the variance. Still, if we want agents to intervene and steer the conversation away from an impasse, they must understand how expressions unfold over time. Prior research on disputes suggests that disputants often reach an impasse because anger provokes anger, leading to escalatory spirals and hardened positions~\cite{pruitt2007conflict}. Thus, we examine whether automatic emotion recognition reveals this pattern.

We examine how disputants express emotion turn-by-turn --- as annotated by GPT4o --- and how each side responds to the other. We do this separately for disputes that end in resolution or impasse. We focus on \textit{anger}, given its recognized role in disputes, but also compassion, as some research highlights its role in shaping outcomes~\cite{Allred97compassion}. Fig.~\ref{fig:emotiontime} illustrates the results.
At the beginning of the dispute, buyers express more anger than sellers, regardless of whether the conversation ends in resolution or impasse. This may reflect that participants internalize typical customer service conversational norms where sellers are expected to express positive emotions in conflict situations~\cite{Hochschild2003,EmLaborTang13}.
Emotion trajectories begin to diverge almost immediately after the first round. For disputes that end in an impasse (the red lines), sellers reciprocate the buyer's anger, leading to an escalatory spiral. For disputes ending in resolution (the green lines), sellers apparently ``stick to the script'' and resist reciprocating the buyer's anger. This seems to defuse the buyers' initial anger, and expressions of anger become less frequent as the conversation continues.

Interestingly, the pattern of compassion also highlights the differences between success and failure. For disputes ending in an impasse, sellers begin with little compassion. In contrast, sellers who reach a resolution begin the dispute by expressing more \textit{compassion} for the situation, and buyers reciprocate those expressions. Thus, our automatic analysis highlights a pathway to dispute resolution that has been under-emphasized in the dispute literature.
Together, these patterns suggest that early differences in expressed emotion can build and shape the outcome. This gives hope that algorithms can be developed to recognize these patterns early and intervene.

\section{Benchmarking}
Our analysis focused on GPT4o due to its growing usage in research. We extend our analysis from GPT4o to other current generative language models to compare against human annotators.
Here, we consider a subset of $N=400$ dialogues.

\subsection{Additional LLMs}
We analyze the following additional LLMs:
\begin{itemize}
    \item \textbf{Deepseek V3:} DeepSeek released a 671B parameter model in January 2025, which outperforms GPT4o at some tasks\cite{liu2024deepseek}.
    \item \textbf{Llama3:} Meta released this open-source 405B parameter model in November 2024\cite{grattafiori2024llama}.
    \item \textbf{GPT4o-mini:} OpenAI released this smaller model in July 2024 --- it performs competitively at many tasks compared to larger, more expensive models\cite{openai2024gpt4omini}.
\end{itemize}


\subsection{Ground Truth}
\begin{figure}[t] 
\centering
\resizebox{\linewidth}{!}{%
    \includegraphics{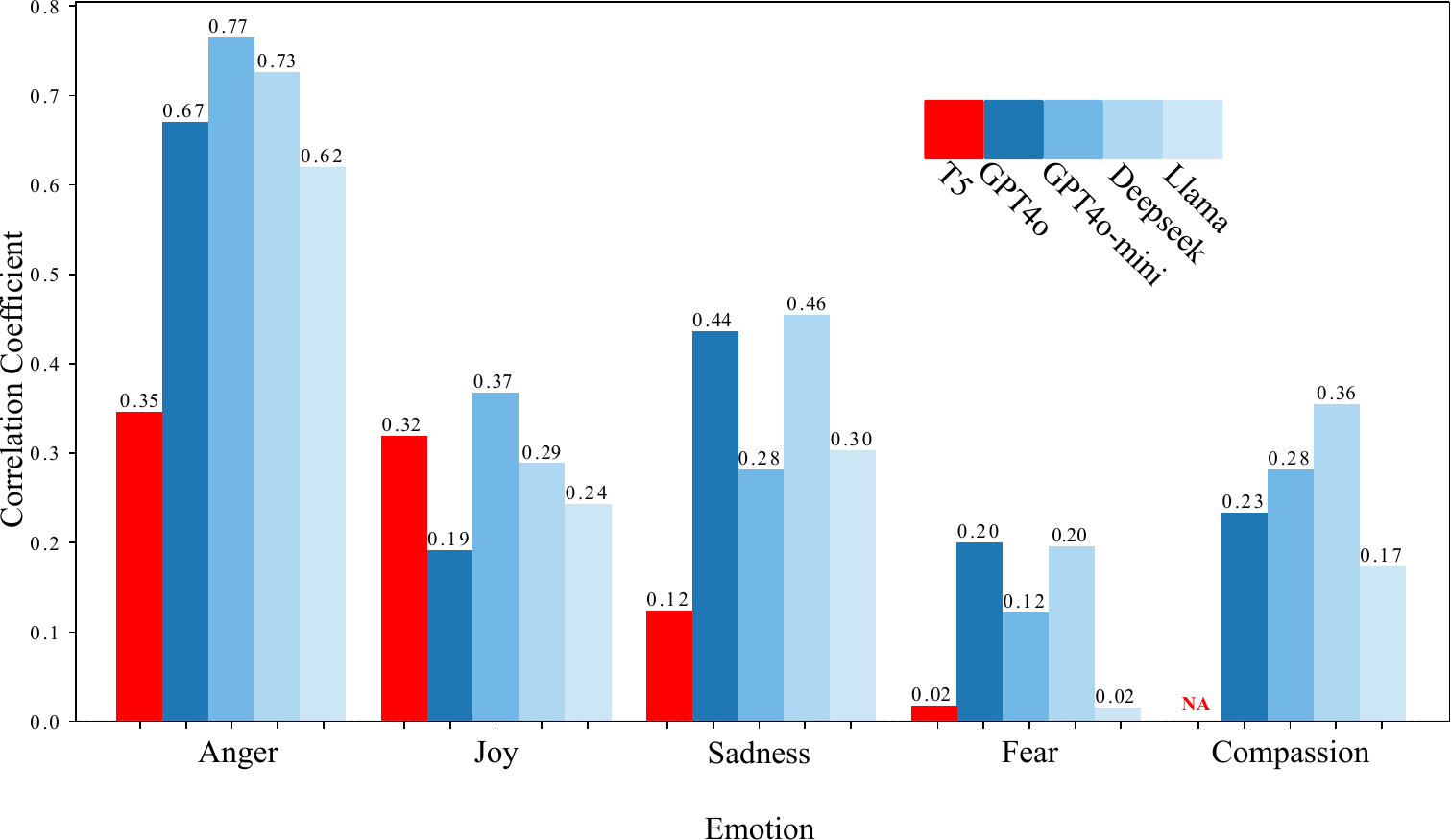} 
}
\caption{Emotion intensity correlations with human annotations --- LLMs better match human annotations.}
\label{fig:corr}
\end{figure}

Previously, we looked at correlations between emotion predictions and self-reported frustration. We expand on this by collecting third-party human emotion intensity annotations ($N=336$) from Prolific (US annotators) on 100 randomly sampled utterances to evaluate T5 and the aforementioned LLMs --- each sampled utterance received at least two annotations from different annotators. The annotators' task resembled that of the models in that they read a highlighted utterance, read the dialogue history, and allocated weight over the considered emotions (\textit{anger}, \textit{joy}, \textit{sadness}, \textit{fear}, \textit{compassion}, and \textit{neutral}) such that they summed to one. A correlational analysis allows us to gauge how well each model matches the average human. Fig.~\ref{fig:corr} illustrates these correlations. Generally, T5 (we map its \textit{love} predictions to \textit{compassion}) performs relatively poorly against the LLMs.

\subsection{Predicting Subjective Outcome}
\begin{figure}[t] 
\centering
\resizebox{\linewidth}{!}{%
    \includegraphics{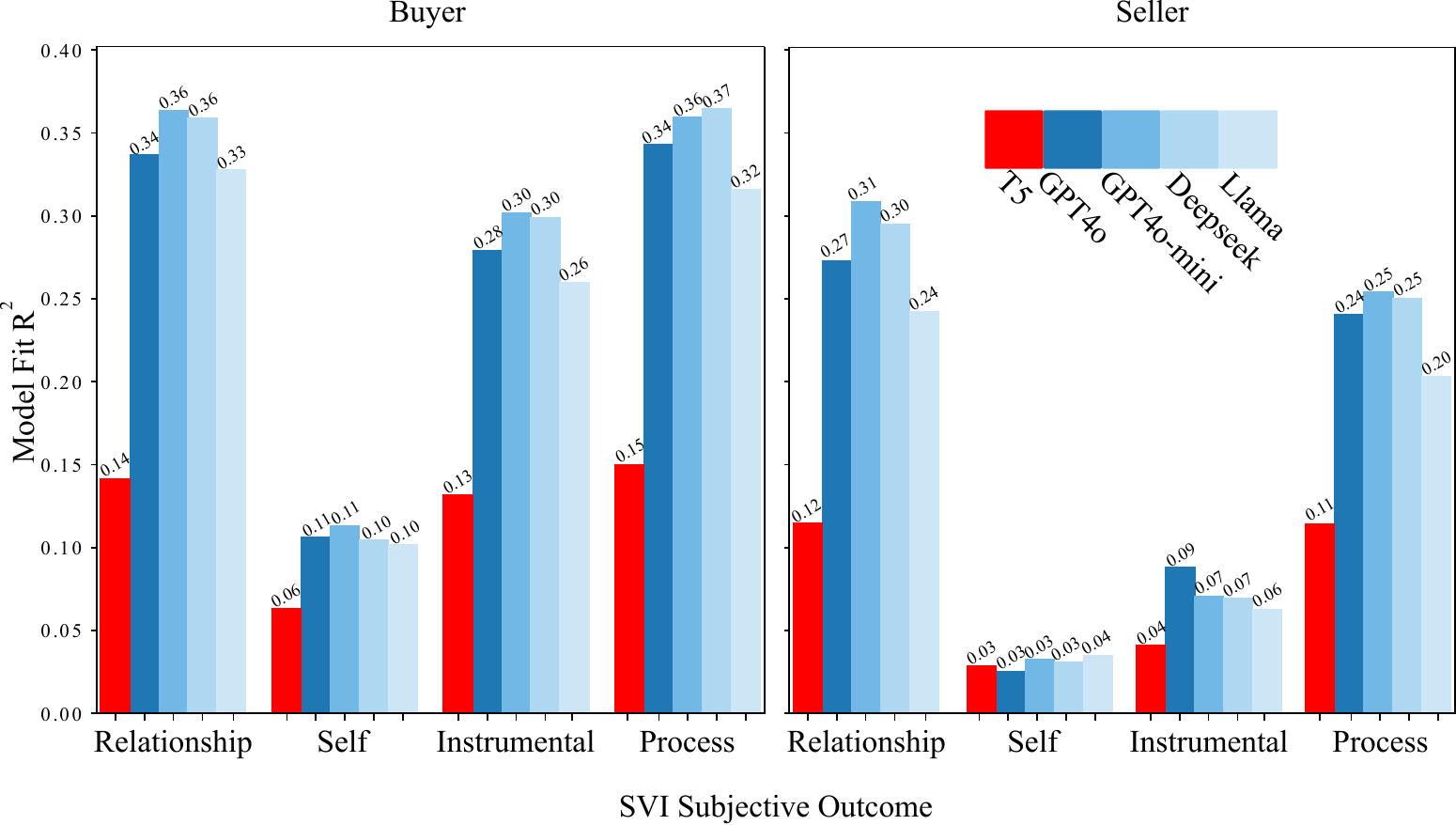} 
}
\caption{Fit of emotion labels predicting subjective outcome --- LLMs explain more variance in subjective outcomes.} 
\label{fig:variousfit}
\end{figure}

As we did in Section~\ref{sec:em-comparison}, we probe these models' abilities to generate emotion labels useful in downstream tasks, such as predicting subjective outcomes. Fig.~\ref{fig:variousfit} demonstrates these models' differing abilities to generate emotion labels useful in this task. Again, we see T5 performs notably worse than the four LLMs at creating useful annotations --- further, the LLM annotations hold quite a bit of power, in some cases explaining about 35\% of the variance from emotion alone.

\section{Discussion}
This study analyzed emotional expressions in a large corpus of buyer-seller disputes. We find LLMs are quite effective at recognizing emotional expressions in dispute dialogues, and far more effective than T5 models used in recent studies. Remarkably, we find that quite of bit of the variance in the subjective outcome of a dispute can be explained from emotional expressions alone, ignoring the actual content of the dialogue.  
For example, simple regression models trained on automatic emotion labels explain, in some cases, thirty to forty percent of the variance in how participants viewed the process and their relationship with their partner.


Analysis also reveals how emotions shape disputes over time. Fig. \ref{fig:anger} illustrates how disputes can escalate into failure if one partner reciprocates anger with anger --- i.e., an escalatory spiral. Buyers begin the dispute, expressing far more anger than their partners. Both sides become increasingly angry when sellers respond angrily, and the dispute ends in an impasse. When sellers maintain their calm, the anger of buyers dissipates, and disputants tend to resolve the dispute. This reinforces findings on escalatory spirals in the social science literature~\cite{pruitt2007conflict}. Interestingly, we also find evidence that spirals of compassion might reverse these effects (see Fig.~\ref{fig:compassion}). Together, our findings suggest agents could intervene early to encourage participants to avoid costly escalation (see~\cite{gelfandLLM24,murawski2024negotiage}).

Our analysis provides prompt engineering insight that might improve text emotion recognition. As highlighted in Table \ref{tab:r2_results}, simply moving from T5 to GPT substantially increased predictive power. However, the results draw attention to other factors that improved the results. For example, including a few labeled examples (in-context learning) aided prediction. Including prior dialogue history also leads to greater performance. Finally, the original T5 model did not allow utterances to be labeled as \textit{neutral}, and it used \textit{love} rather than \textit{compassion} --- changing these labels also enhanced performance.

Lastly, we benchmarked several LLMs against human annotations and assessed how well their predictions support modeling subjective outcomes. The four LLMs generally outperform T5 and exhibit varying relative performance across different tasks --- e.g., correlating with human ground truth and explaining subjective outcomes. 

\section{Conclusion \& Future Work}
There are clear individual and societal benefits to agents understanding and helping resolve emotional disputes. Unresolved disputes can promote social divisions, erode trust in groups and institutions, and cause mental and emotional distress to the parties involved. 
Future work may wish to analyze how gender and culture interact to shape automatically recognized emotion. However, in our results, we do not find much evidence of interactions between those demographic factors, as one might expect if there existed a systemic issue of misclassification. Nonetheless, this remains a salient question going forward, especially as we work toward our ultimate goal of AI managing human disputes. 

If AI-driven systems could recognize the emotional dynamics of how disputes arise and escalate, they have the potential to manage these negative costs. 
Ultimately, we will leverage the findings here to create agents that can effectively mediate human disputes. We previously published preliminary findings on the abilities of LLMs in mediation \cite{hale2025ai} — finding the model 
picked up on salient dialogue features to intervene more so in disputes resulting in impasse, and in those with high frustration. Further, we find
crowd-source workers rated AI-generated interventions higher compared to novices. Additionally, we will move beyond \textit{just} emotion and explore methods of annotating the semantic content of conflict dialogues, following pertinent research~\cite{kwon2024llmseffectivenegotiatorssystematic,friedman2024application}. 

\section{Ethical Impact Statement}
Several ethical concerns must be considered as this work proceeds.  Prior work has suggested that LLMs internalize societal and cultural stereotypes, and any attempts to deploy such technology must carefully consider the risks of misrepresenting or misinterpreting perceived emotion. Quite possibly, certain groups might be labeled as ``angry'' and treated unfairly by automated methods. This is particularly a concern in cross-cultural contexts where emotions might hold different connotations and serve different functions~\cite{havaldar2023multilingual}. 
Finally, though we focused on dispute \textit{resolution}, our findings could inform techniques that intentionally escalate conflicts.  This emphasizes the importance of increased research in conflict processes and techniques, how they might be deployed, and their consequence for society.

 \textcolor{black}{Secondly, we acknowledge the limited purview of our work and the uncertainty of whether these results could extend outside the realm of dispute resolution. However, we wish to reiterate that \textit{this} kind of generalizability lies outside the scope of this work; rather, we intended to demonstrate the feasibility of using emotion annotations from LLMs in the understudied domain of dispute resolution. Here, emotion comes to the forefront compared to the more commonly adopted — at least in computer and affective science — task of deal-making. We find, in some cases, linear models can explain almost 40\% of the variance in subjective outcome considering just these emotion labels, supporting the importance of affect in this domain. }

This work was approved by the University of Southern California's institutional review board (UP-24-01183-AM001).



\section*{Acknowledgment}
  This work is supported by the U.S. Government including the Air Force Office of Scientific Research (grant FA9550-23-1-0320), the Army Research Office under agreement W911NF-25-2-0040, and the National Science Foundation (grant 2150187). The views and conclusions contained in this document are those of the authors and should not be interpreted as representing the official policies, either expressed or implied, of the Army Research Office or the U.S. Government. The U.S. Government is authorized to reproduce and distribute reprints for Government purposes notwithstanding any copyright notation herein.

\bibliographystyle{IEEEtran}
\bibliography{sample}

\end{document}